\documentclass[lettersize,journal]{IEEEtran}
\usepackage{amsmath,amsfonts}
\usepackage{algorithmic}
\usepackage{algorithm}
\usepackage{array}
\usepackage[caption=false,font=normalsize,labelfont=sf,textfont=sf]{subfig}
\usepackage{textcomp}
\usepackage{stfloats}
\usepackage{url}
\usepackage{verbatim}
\usepackage{graphicx}
\usepackage{cite}
\usepackage{multirow}
\usepackage{xcolor}
\usepackage{orcidlink}
\hypersetup{hidelinks}
\begin{document}

\title{SSFMamba: Learning Symmetry-driven Spatial-Frequency Modeling for Physically Consistent 3D Medical Image Segmentation}

\author{
    Bo Zhang$^{\orcidlink{0000-0002-1210-8735}}$,~\IEEEmembership{Member,~IEEE,}
    Yifan Zhang$^{\orcidlink{0009-0003-0071-6711}}$, 
    Shuo Yan$^{\orcidlink{0000-0003-1092-455X}}$, 
    Yu Bai$^{\orcidlink{0000-0001-9923-7427}}$, 
    Zheng Zhang$^{\orcidlink{0000-0002-8207-6089}}$,  
    Wu Liu$^{\orcidlink{0000-0003-1633-7575}}$,~\IEEEmembership{Member,~IEEE,}
    Xiuzhuang Zhou$^{\orcidlink{0000-0001-8838-873X}}$,~\IEEEmembership{Member,~IEEE,} 
    Wendong Wang$^{\orcidlink{0000-0002-6418-8087}}$,~\IEEEmembership{Member,~IEEE}
\thanks{Manuscript received March 1, 2026; revised August 26, 2026; accepted October 22, 2026. 
This work was supported in part by the National Natural Science Foundation of China (No. 62676045, 62606163 and 62576049), the Noncommunicable Chronic Diseases-National Science and Technology Major Project (No. 2025ZD0546100 and 2025ZD0546102), and the Beijing Natural Science Foundation (No. L2602064).}
\thanks{Bo Zhang, Yifan Zhang, Shuo Yan, Yu Bai and Wendong Wang are with the State Key Laboratory of Networking and Switching Technology, School of Computer Science (National Pilot Software Engineering School), Beijing University of Posts and Telecommunications. E-mail: \{zbo, zy1f4n, shuoyan, by, wdwang\}@bupt.edu.cn.}
\thanks{Zheng Zhang and Xiuzhuang Zhou are with the School of Intelligent Engineering and Automation, Beijing University of Posts and Telecommunications. E-mail: \{zhangzheng, xiuzhuang.zhou\}@bupt.edu.cn.}
\thanks{Wu Liu is with the School of Information Science and Technology, University of Science and Technology of China. E-mail: liuwu@ustc.edu.cn.}
\thanks{Corresponding author: Zheng Zhang.}
}

\markboth{IEEE Transactions on Image Processing,~Vol.~99, No.~99, October~2026}%
{Bo \MakeLowercase{\textit{et al.}}: SSFMamba: Learning Symmetry-driven Spatial-Frequency Modeling}

\IEEEpubid{0000--0000/00\$00.00~\copyright~2026 IEEE}

\maketitle

\begin{abstract}
Accurate 3D medical image segmentation requires a delicate balance between fine-grained local details and global contextual understanding. While spatial-domain models often struggle with long-range dependencies, existing frequency-based approaches frequently overlook intrinsic spectral properties such as Hermitian symmetry, leading to suboptimal feature integration. In this paper, we propose SSFMamba, a \textbf{Mamba}-based \textbf{S}ymmetry-driven \textbf{S}patial-\textbf{F}requency fusion framework tailored for 3D medical imaging. Our architecture employs a complementary dual-branch design: the spatial branch models voxel sequences to preserve intricate anatomical textures, while the frequency branch operates on spectral components, each of which aggregates information from the entire volume, to capture global context. A core innovation is the 3D Multi-Directional Scanning Mechanism (MDSM), which integrates Hermitian symmetry with the causal nature of State Space Models (SSMs) to enable direction-aware global modeling. Crucially, by shifting the modeling focus to frequency-domain spectral components, SSFMamba captures the underlying structural characteristics of anatomical tissues. This leads to a highly adaptable framework that excels in both MRI and CT applications, regardless of the significant variations in intensity distributions. Extensive evaluations on the BraTS2020, BraTS2023, and BTCV datasets demonstrate that SSFMamba consistently outperforms state-of-the-art methods. Notably, our approach achieves exceptional performance on low-contrast organs such as the pancreas (81.97\% Dice), underscoring its potential as a unified and physically consistent perception framework for diverse 3D clinical applications.
\end{abstract}


\section{Introduction}
\label{sec:intro}
\IEEEPARstart{T}{he} aim of 3D medical image segmentation is to accurately delineate the complex anatomical structures of organs or tumors, which requires not only fine-grained modeling of local details but also a holistic understanding of spatial relationships. 
Recent studies that rely solely on spatial-domain information are effective at capturing local features, but they often fall short in modeling long-range dependencies, particularly in high-resolution, large-volume 3D medical images. This limitation leads to challenges such as inaccurate boundary delineation, difficulty in handling complex or spatially extended structures, and a tendency to overfit to local patterns. For example, CNN models \cite{lecun1995convolutional} can effectively capture local features, but are inherently limited when modeling long-range dependencies. While Transformer models \cite{vaswani2017attention} excel at capturing global context, their quadratic computational complexity introduces a significant scaling bottleneck for 3D medical volumes. This overhead becomes computationally prohibitive in high-resolution or multimodal scenarios, where the integration of heterogeneous voxel data substantially increases the resource burden. Recently, Mamba models \cite{gu2023mamba} have demonstrated strong capabilities in modeling long-range dependencies by leveraging a state-space formulation combined with a selective computation mechanism. Mamba reduces the computational complexity to a linear relationship with sequence length, making it well-suited for medical imaging applications. However, Mamba is sensitive to the ordering of input sequences, and the conversion of spatial structures into one-dimensional sequences presents challenges for preserving and effectively modeling spatial relationships. 

\begin{figure}[!tb]
    \centering
    \includegraphics[width=0.98\linewidth]{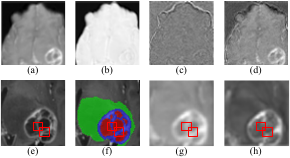}
    \vspace{-3mm}
    \caption{(a) Original image and feature maps from (b) the spatial domain, (c) the frequency domain, and (d) the fused multi-domain representation; (e) original image, (f) annotation, (g) our feature map, and (h) SegMamba's feature map. The spatial domain provides comprehensive representations but exhibits blurred boundaries; the frequency domain provides clear boundaries and high-contrast details, and the fused features can effectively emphasize the tumor region and its peripheral contours by integrating both representations.
    }
    \label{fig:comparison}
    \vspace{-4mm}
\end{figure}

\IEEEpubidadjcol

To overcome the restricted receptive field and lack of global awareness in spatial-domain models, frequency domain analysis has garnered increasing attention. Compared to natural images, the utility of frequency-domain features is even more pronounced in medical imaging. 
For instance, magnetic resonance imaging (MRI) data are inherently acquired in the frequency domain (k-space)~\cite{plewes2012physics}, aligning frequency-based representations more closely with the physical foundation of medical images. 
The frequency domain decomposes spatial variations in an image into distinct frequency components: low frequencies represent the overall structure and background, while high frequencies correspond to edges and fine details. This decomposition provides explicit access to global structures and boundary information, which is complementary to spatial-domain representations. 
For example, incorporating frequency-domain information has proven effective in enhancing a model's global perceptual capability in image fusion and enhancement applications~\cite{zhao2023cddfuse, zou2024wave}.
As illustrated in Fig.~\ref{fig:comparison}, the locally enlarged details in (e–h) demonstrate that our method preserves edge details more effectively compared to feature maps that rely solely on spatial-domain information.

However, researchers often use naive transformations such as $1\times1$ convolution to extract frequency-domain features \cite{zou2024freqmamba, chang2024net}, but this approach fails to fully leverage the unique characteristics of frequency data. 
Compared to the spatial domain, the frequency domain exhibits distinctive properties such as Hermitian symmetry, compact energy concentration, and global receptive field representation. 
Notably, while Hermitian symmetry has been extensively utilized in traditional compression algorithms \cite{rhee2022lc, velisavljevic2007space, xu2020learning}, its utility in high-fidelity image processing remains largely untapped, representing a missed opportunity for further performance improvements. 

These properties offer the potential for more efficient encoding of structural and contextual information. 
Meanwhile, the frequency domain also introduces nontrivial challenges: its semantic representations are less intuitive, and its data distributions fundamentally differ from those in the spatial domain. 
Consequently, naive integration with spatial features may dilute or obscure the complementary strengths of frequency-based representations. 

To address the aforementioned challenges, we propose Symmetry-driven Spatial-Frequency Feature Fusion Mamba (SSFMamba), as illustrated in Fig.~\ref{fig:SSFMamba}. We investigate the significance of frequency-domain representations in medical imaging and propose a dual-branch feature fusion network to separately extract spatial and frequency-domain features. 
SSFMamba employs the Mamba module to extract features from frequency-domain representations and fully harnesses its global context extraction capabilities and the synergistic advantages of frequency-domain structural information to enhance global feature modeling. 
To this end, we optimize the architectural design so that Mamba can efficiently process and integrate information from both spatial and frequency domains, achieving deep coupling and synergistic enhancement of both global context and local details. 
Since Mamba is sensitive to the order of its input sequences, as noted above, by leveraging the Hermitian symmetry of frequency-domain data for sequence modeling, we significantly enhance its global modeling capability. 
Beyond global modeling, these spectral properties also benefit cross-modality consistency. While spatial intensities—such as Hounsfield Units in CT and relaxation signals in MRI—are governed by distinct physical imaging protocols, the underlying anatomical boundaries and organ envelopes exhibit a high degree of spectral congruence. By shifting the modeling focus from intensity-dependent spatial voxels to frequency-domain spectral components, SSFMamba captures modality-invariant ``structural footprints'' of human anatomy and alleviates the intensity-based domain shifts inherent in spatial representations. Consequently, SSFMamba establishes a unified and physically consistent perception framework that generalizes across diverse clinical tasks, from brain tumor segmentation in MRI to multi-organ delineation in abdominal CT. 
Our main contributions are as follows:
\begin{itemize}
\item We propose a physically consistent spatial-frequency perception framework that complements modality-dependent spatial representations with structural cues in the frequency domain, enabling a unified architecture to effectively handle heterogeneous 3D MRI and CT segmentation tasks.

\item We introduce the Frequency Mamba, which leverages Hermitian symmetry to enhance global modeling with causal State Space Models (SSMs).
This enables adaptive spectral feature extraction, allowing the model to dynamically emphasize critical anatomical signatures.

\item We develop a 3D Multi-Directional Scanning Mechanism (MDSM) that serializes the spectrum along complementary in-slice, cross-slice, and 3D local orders. It strengthens global spectral modeling while retaining planar, through-plane, and volumetric neighborhood structures, thereby improving boundary delineation in large-volume 3D scans.

\item We achieve state-of-the-art performance on the multimodal MRI BraTS datasets and the CT-based BTCV dataset, notably reaching 81.97\% Dice on the pancreas. This validates the framework's robust generalization and clinical potential across heterogeneous modalities and complex anatomical regions.
\end{itemize}

\section{Related Work}

\subsection{3D Medical Image Segmentation} 
Many studies focus on striking a better balance between global modeling and detail recovery, thereby providing more accurate segmentation results for practical clinical applications. In early research, CNN-based methods primarily focused on extracting local features, which significantly improved segmentation performance within the U-Net \cite{ronneberger2015u} architecture. Built upon the U-Net architecture, UX-Net \cite{lee2023uxnet} enhances volumetric medical image segmentation by employing U-shaped convolutional blocks with large receptive fields, enabling efficient multi-scale context modeling while maintaining a low computational cost. 
Moreover, MedNeXt \cite{roy2023mednext} leverages large-kernel depthwise convolutions to expand the receptive field of CNNs, effectively capturing long-range spatial dependencies with a lightweight architecture for 3D medical image segmentation.
However, these approaches are limited by the receptive fields of CNNs, making it challenging to effectively model global information. With the successful application of Transformers in computer vision, an increasing number of studies have begun exploring hybrid architectures that combine Transformers with CNNs, such as UNETR \cite{hatamizadeh2022unetr}, TransUNet \cite{chen2024transunet} and SwinUNETR \cite{hatamizadeh2022swinunetr}. TransUNet integrates Transformer modules into the encoder to model long-range dependencies, effectively combining global contextual representations with convolutional inductive biases for improved medical image segmentation. SwinUNETR employs a hierarchical Swin Transformer as the encoder, leveraging shifted-window self-attention to achieve efficient multi-scale representation learning and enhanced global context modeling in 3D medical image segmentation. By leveraging the excellent global modeling capabilities of Transformers alongside the local modeling strengths of CNNs \cite{tang2022self, li2022cats}, segmentation performance has been further improved. However, one significant drawback of Transformers is their quadratically increasing computational complexity, which makes scaling to 3D images with a large number of voxels challenging. 

In recent years, the emergence of Mamba has brought new opportunities, as it proves to be more efficient than traditional Transformer networks when dealing with large-scale 3D images \cite{liu2024swin, liu2024vmamba}. However, its potential in 3D vision tasks remains far from fully explored.
In this work, we propose a module that leverages FFT in conjunction with Mamba to extract global frequency information and fuse it with spatial-domain features. This Multi-domain Information Fusion (MDIF) module further enhances segmentation performance.

\subsection{Multi-domain Information Fusion} 
Traditional image processing models predominantly rely on spatial-domain features due to the intuitive nature of pixel values and their well-defined spatial relationships, which have led to significant achievements across various vision tasks \cite{liu2021swin, wenxuan2021transbts, xie2024csfwinformer}. However, frequency-domain information provides unique advantages in capturing global patterns by encoding frequency properties that are often challenging to extract using spatial-domain methods alone. Frequency-based representations enhance global structure perception, thereby improving model performance and robustness.

The Fourier transform serves as a crucial tool for converting spatial-domain information into the frequency domain, enabling the extraction of global statistical properties that have been widely applied in numerous computer vision tasks, such as image denoising~\cite{sheng2022frequency,kong2023efficient,kaur2023complete}, segmentation~\cite{li2023global, qiong2025medical}, and image restoration and enhancement \cite{huang2022deep, pratt2017fcnn, zhou2022adaptively, liu2023reducing, zou2024freqmamba,dudhane2023burstormer}. Given the complementary strengths of spatial and frequency-domain representations, many existing approaches integrate both to achieve a balance between preserving fine-grained local details and maintaining a coherent global context, thereby enhancing overall image processing effectiveness.  
For example, SFFNet \cite{yang2024sffnet} adopts a dual-branch design that integrates wavelet-based frequency-domain features with spatial-domain features for remote sensing segmentation. Similarly, frequency-domain image deraining techniques, such as FreqMamba \cite{zou2024freqmamba}, mitigate global degradation by leveraging the strong correlation between raindrop effects and the Fourier amplitude spectrum.  
It is worth noting that EM-Net~\cite{chang2024net} also incorporates frequency-domain information into a Mamba-based segmentation network. However, EM-Net essentially relies on Mamba for feature extraction in the spatial domain after transforming frequency features back, without exploiting the modeling capability of state-space models in the frequency domain.

Despite notable progress in incorporating frequency-domain information \cite{wang2023fremae}, its full potential remains underutilized. 
In general, spatial-domain representations focus on local contextual relationships and fine-grained details, whereas frequency-domain representations emphasize global structures and pattern properties. The integration of these two types of information creates a complementary effect, preserving intricate local details (such as edges and textures) while enhancing global perception (such as regional contrast and object contours). 

\subsection{State Space Models}
Recent advances in structured SSMs \cite{gu2021efficiently, smith2022simplified} have garnered significant attention from both academia and industry due to their computational efficiency in long-sequence processing. Unlike traditional Transformer architectures with quadratic computational complexity $\mathcal{O}(L^2)$, SSMs achieve linear complexity $\mathcal{O}(L)$ through hidden state recursion mechanisms, reducing resource consumption in long-context scenarios. The seminal work S4 (Structured State Space Sequence Model) \cite{gu2021efficiently} established the foundation for modern deep learning in long-sequence modeling. 

Notably, Mamba \cite{gu2023mamba}, an evolved version of S4, has overcome the expressiveness limitations inherent in conventional SSMs. Empirical studies demonstrate that Mamba not only matches or surpasses Transformer performance in tasks such as language modeling and genomic analysis, but also rigorously maintains linear scalability with sequence length, offering a novel technical pathway for low-resource long-sequence inference. Mamba has transcended its original application in natural language processing, achieving remarkable progress in tasks such as image segmentation and fusion, as well as in other domains including medical imaging analysis and computer vision \cite{liu2024vision, yue2024medmamba, lianghui2024vision, bao2025vision}. MedMamba \cite{yue2024medmamba} introduces Mamba-based state space models into medical image classification, efficiently modeling long-range dependencies with linear complexity.

\begin{figure*}[!tbp]
  \centering
  \includegraphics[width=0.95\textwidth, angle=0]{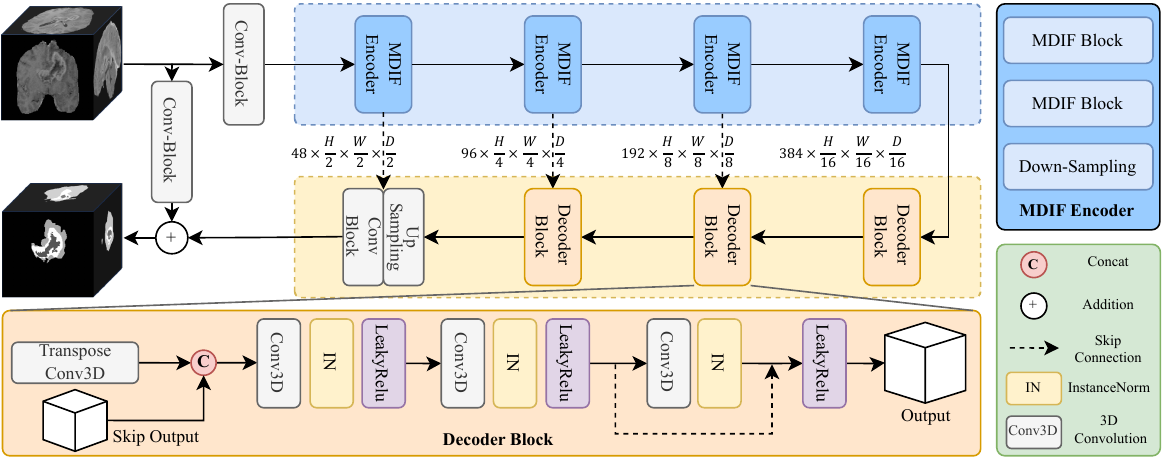}
    \vspace{-2mm}
  \caption{The overall architecture of the proposed SSFMamba. The encoder utilizes convolutional layers and multiple Multi-domain Information Fusion (MDIF) Encoders to extract multi-scale features. Each MDIF Encoder consists of two MDIF Blocks and a downsampling layer, designed to capture and integrate multi-domain information. Notably, the MDIF Blocks are based on Mamba Blocks. In the decoder, transposed convolutions upsample the feature maps, which are then concatenated with corresponding skip connection feature maps from the encoder. Subsequent convolutional layers perform feature fusion to restore high-resolution representations. Additionally, residual connections are incorporated to facilitate effective deep network training.
  }
  \label{fig:SSFMamba}
\end{figure*}

\section{Method}

\subsection{Preliminaries}

\subsubsection{Frequency Analysis}\label{sec:freq}

Medical images typically exhibit complex anatomical structures, noise interference, and intensity variations between tissues. The frequency domain is of great significance to global feature extraction, noise suppression and enhancement, and multimodal data fusion. By providing a global perspective for medical image interpretation via mathematical transformations (e.g., Fourier transform, wavelet transform), the frequency domain compensates for the inherent limitations of conventional spatial-domain methodologies. 
The 3D Fast Fourier Transform (3D FFT)~\cite{lucarini2021fft, schneider2021review} is a computational algorithm that decomposes a three-dimensional image into a set of frequency components. For instance, consider a 3D image with dimensions $H\times W\times D$. The mathematical formulation of its 3D Discrete Fourier Transform (3D DFT) can be expressed as follows:

{\small
\begin{equation}
F(u, v, w) = \sum_{x=0}^{H-1} \sum_{y=0}^{W-1} \sum_{z=0}^{D-1} 
f(x, y, z) e^{-j2\pi \left( \frac{ux}{H} + \frac{vy}{W} + \frac{wz}{D} \right)},
\end{equation}
}

where \(x\), \(y\) and \(z\) denote the spatial indices, and \(u\), \(v\) and \(w\) denote the frequency indices along the height, width, and depth dimensions, respectively. The function \(f\) represents the original 3D signal in the spatial domain. 
This process enables the synergistic representation of both global structural properties and local fine-grained details through their frequency-domain features. 
Specifically, when the original 3D image consists of real-valued data, its representation in the frequency domain exhibits Hermitian symmetry \cite{walker2017fast}. Namely, the complex conjugate of a frequency component is equal to the value at its symmetric position across the origin:

{\small
\begin{equation}
F(-u, -v, -w) = \overline{F(u, v, w)}
\label{eq:hermitian}
\end{equation}
}

This symmetry implies that the 3D frequency tensor contains redundant but complementary information pairs. In particular, the magnitude spectrum satisfies $|F(-u,-v,-w)| = |F(u,v,w)|$. With the standard (unshifted) FFT layout, the conjugate counterpart of index $k$ along a dimension of length $N$ is $(N-k) \bmod N$. Consequently, when the magnitude spectrum is flattened in raster order, the resulting sequence is approximately palindromic, i.e., it is mirror-symmetric except for the entries on the zero-frequency planes. Since the convolutional block and normalization applied before the Frequency Mamba are point-wise, this symmetry is preserved in its input sequence. 
A single causal forward scan therefore visits each spectral component and, later, its mirrored counterpart, so that the second half of the sequence is processed with a hidden state that has already summarized the first half in the opposite order. This provides both forward and reverse views of the spectrum within one scan and alleviates the directional bias of causal SSMs without additional backward scanning. Our MDSM is built upon this property, as detailed in Section~\ref{sec:mdsm}.

\subsubsection{State Space Models} 
The selective SSM utilized in Mamba builds upon the Structured State Space Model (S4) and extends traditional state space models by enhancing expressive power through discretizing continuous systems and incorporating input-dependent parameters. It is specifically designed for long-sequence modeling, with its core mathematical formulation as follows:

{\small
\begin{equation}
\begin{gathered}
h'(t) = \mathbf{A} h(t) + \mathbf{B} x(t),\\
y(t) = \mathbf{C} h(t) + \mathbf{D} x(t).
\end{gathered}
\end{equation}
}

Traditional SSMs rely on fixed matrices ($\mathbf{A}$, $\mathbf{B}$, $\mathbf{C}$), limiting their capacity to capture context-sensitive dynamics. Mamba addresses this by introducing a selective mechanism, where $\mathbf{B}$ and $\mathbf{C}$ are dynamically generated via input-conditioned linear projections, enabling adaptive feature representation. Matrix $\mathbf{A}$ is initialized using the HiPPO framework to enhance long-term memory and efficient compression of historical information. This design allows the model to selectively emphasize salient elements while suppressing irrelevant content.

Given Mamba's sensitivity to input sequence order and its limited capacity for local spatial modeling, we leverage the structural properties of frequency-domain information—symmetry and directional distribution—to design a multi-directional scanning mechanism tailored for 3D frequency-domain information. This strategy effectively balances global context modeling with local feature extraction.

\begin{figure*}[!tbp]
  \centering
  \includegraphics[width=0.9\textwidth, angle=0]{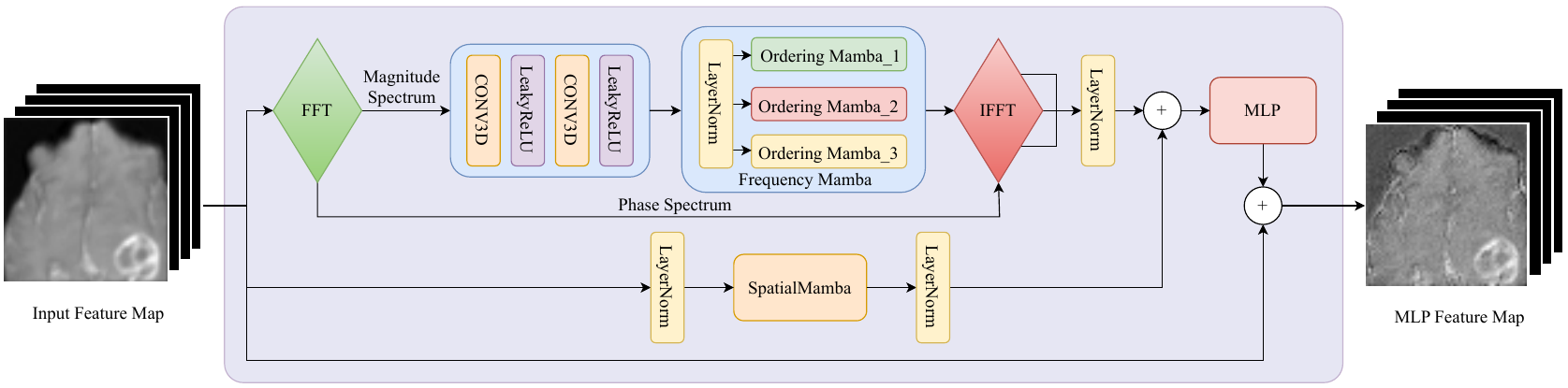}
    \vspace{-2mm}
  \caption{The detailed architecture of the proposed MDIF Block. This block consists of two branches: one models voxel sequences in the spatial domain to preserve fine-grained local details, while the other models the magnitude spectrum, in which each token corresponds to a global frequency component, to capture global context. The features from both branches are then fused using an MLP module and combined with the input feature maps to generate the output feature maps.
  }
  \label{fig:MDIFBlock}
\end{figure*}

\begin{algorithm}[!tb]
\caption{MDIF Block (SSFMambaLayer)}
\begin{algorithmic}[1]
\REQUIRE Input tensor $x$ of shape $(B, C, H, W, D)$
\ENSURE Output tensor $out$
\STATE $x_{flat} = \text{reshape}(x, (B, C, HWD))^\top$ \COMMENT{Flatten spatial dims}
\STATE $x_{norm} = \mathrm{LayerNorm}(x_{flat})$
\STATE $x_{sp} = \mathrm{Mamba_{spatial}}(x_{norm})$
\STATE $x_{sp} = \text{reshape}(x_{sp}^\top, (B, C, H, W, D))$
\STATE $x_{fft} = \mathrm{FFT}( \text{reshape}(x_{norm}^\top, (B, C, H, W, D)) )$
\STATE $mag = |x_{fft}|$, \quad $pha = \angle(x_{fft})$
\STATE $mag = \mathrm{ConvBlock}(mag)$ \COMMENT{Point-wise conv}
\STATE $mag = \mathrm{reshape\&norm}(mag)$
\STATE $s_{cross} = \mathrm{Flatten}_{(H,W,D)}(mag)$ \COMMENT{Cross-slice}
\STATE $s_{in} = \mathrm{Flatten}_{(D,W,H)}(mag)$ \COMMENT{In-slice}
\STATE $s_{local} = \mathrm{Patches}_{4\times4\times4}(mag)$ \COMMENT{3D local}
\FOR{$k \in \{cross, in, local\}$}
\STATE $s_{k} = \mathrm{Mamba}_{k}(s_{k})$
\STATE Reshape $s_{k}$ back to $(B,C,H,W,D)$
\STATE $x_{k} = \mathrm{InverseFFT}(s_{k}, pha)$
\ENDFOR
\STATE Normalize $x_{cross}, x_{in}, x_{local}, x_{sp}$
\STATE $x_{sum} = x_{sp} + x_{cross} + x_{in} + x_{local}$
\STATE $x_{mlp} = \text{MLPChannel}(x_{sum})$
\STATE $out = x + x_{mlp}$
\STATE \textbf{return} $out$
\end{algorithmic}
\label{alg:SSFMambaLayer}
\end{algorithm}

\subsection{Architecture} 
The overall architecture of our model is illustrated in Fig.~\ref{fig:SSFMamba}. The proposed model architecture adopts an encoder-decoder framework, where the encoder incorporates MDIF Blocks. Each MDIF Block employs a dual-branch design to extract features from images at multiple scales, fusing spatial and frequency features to capture both global and local contextual information at each scale, while the MDSM is introduced to further strengthen global dependency modeling and improve overall performance. In the decoder, transposed convolutions are employed for upsampling, while skip connections and residual connections are integrated to enhance feature fusion. 
In summary, when a 3D medical image is fed into our model, it is processed through both the spatial and frequency-domain branches for feature extraction. The resulting features are then integrated within the fusion module to generate the input for the next layer, and the multi-scale encoder outputs are passed to the decoder through skip connections. 
This feature fusion process is repeated hierarchically, and finally, the decoder progressively reconstructs the original image size to produce the final segmentation output. Subsequent sections elaborate on the design principles and implementation details of each sub-module.

\subsubsection{\textbf{Multi-domain Information Fusion Block}} 
As illustrated in Fig.~\ref{fig:MDIFBlock}, the MDIF Block comprises two primary branches designed to effectively integrate information from multiple domains, thereby enhancing model performance. 
The implementation details of the MDIF Block are formalized in Algorithm~\ref{alg:SSFMambaLayer}, which summarizes the computation of both branches and their fusion. 
The two branches operate in parallel on the same layer-normalized input: the spatial branch models voxel sequences directly, while the frequency branch transforms the input into the frequency domain to capture complementary frequency-aware features. Finally, the outputs from both branches are fused to form a unified representation. The detailed design of this module is described in the following paragraphs.

\textbf{Frequency-Domain Branch.} For each frequency-domain branch, the input feature \( F_{i,j} \) (i.e., the input of the $j$-th MDIF Block in the $i$-th encoder stage) is first normalized by layer normalization and then transformed into frequency-domain information \( F_{fre} \) using the Fast Fourier Transform (FFT). Since frequency-domain data are inherently complex numbers, they are decomposed into the magnitude spectrum \( F_{mag} \) and the phase spectrum \( F_{pha} \). The magnitude spectrum is then refined by a point-wise convolutional block. The refined magnitude features are serialized by the MDSM (detailed in Section~\ref{sec:mdsm}) into three directional sequences, namely cross-slice, in-slice, and 3D local orderings, where the 3D local ordering partitions the spectrum into non-overlapping $4\times4\times4$ cubes to encode regional spectral signatures. Each sequence is processed by an individual Mamba to capture global spectral dependencies:

{\small
\begin{equation}
\begin{gathered}
    F_{fre} = \mathrm{FFT}(\mathrm{LayerNorm}(F_{i, j})),\\
    F_{mag} = \mathrm{ABS}(F_{fre}),\\
    F_{pha} = \mathrm{ANGLE}(F_{fre}),\\
    F_m = \mathrm{LR}(\mathrm{ConvBlock}_{1\times1\times1}(\mathrm{LR}(\mathrm{ConvBlock}_{1\times1\times1}(F_{mag})))),\\
    F_1, F_2, F_3 = \mathrm{MDSMBlock}(F_m),\\
    \hat{F}_k = \mathrm{Mamba}_k(F_k), \quad k=1, 2, 3,\\
    F_{fre_k} = \mathrm{LayerNorm}(\mathrm{IFFT}(\hat{F}_k, F_{pha})),
\end{gathered}
\end{equation}
}

where IFFT denotes the inverse FFT. In frequency-domain image processing, the ABS and ANGLE operations are used to extract the magnitude and phase spectra from the complex representation of a frequency-domain image, respectively. Correspondingly, these features are denoted as \(F_{mag}\) (magnitude feature) and \(F_{pha}\) (phase feature). LR denotes the LeakyReLU, and MDSMBlock denotes the MDSM block, which reorders the frequency-domain features into three directional sequences.

\begin{figure*}[!tbp]
  \centering
  \includegraphics[width=0.75\textwidth, angle=0]{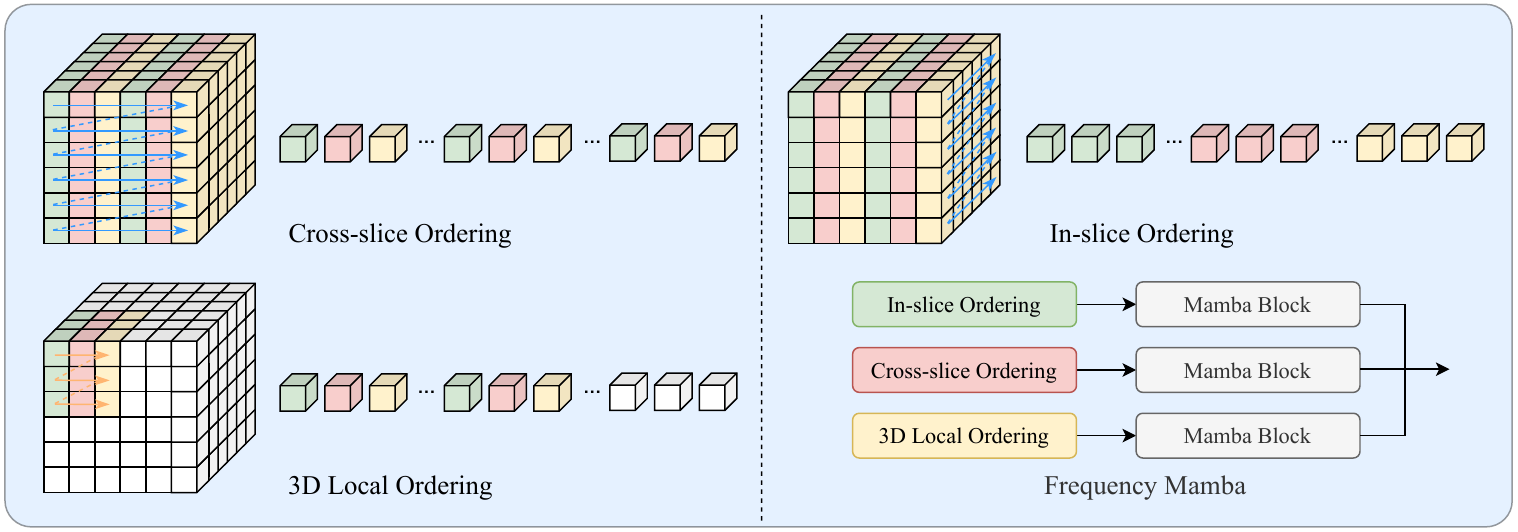}
    \vspace{-2mm}
  \caption{The detailed architecture of the proposed Frequency Mamba with the 3D multi-directional scanning mechanism (MDSM). In the MDSM, features are extracted along different orientations using distinct Mamba modules, and different colors represent different slices. These sequences are then transposed back to their original 3D arrangement. After applying the IFFT, we perform element-wise addition at the same spatial positions.
  }
  \label{fig:MDSM}
\end{figure*}

\begin{algorithm}[!tb]
\caption{SSFMamba Encoder (Dual-Domain features)}
\begin{algorithmic}[1]
\REQUIRE Input tensor $x$
\ENSURE Tuple of feature maps $(f_0, f_1, f_2, f_3)$
\STATE $x = \mathrm{Conv3D}(x;\text{out\_channels}=dims[0], $
\STATE \hspace{1.5em} $
\text{kernel}=7, \text{stride}=2, \text{padding}=3)$
\FOR{$i = 0$ \TO $3$}
    \IF{$i > 0$}
    \STATE $x = \text{InstanceNorm3D}(x)$
    \STATE $x = \mathrm{Conv3D}(x; \text{out\_channels}=dims[i], $
    \STATE \hspace{1.5em} $\text{kernel}=2, \text{stride}=2)$
    \ENDIF
    \STATE \text{Repeat $depths[i]$ times: } $x = \text{SSFMambaLayer}(x)$
    \STATE $x_{out} = \text{InstanceNorm3D}(x)$
    \STATE $f_i = \text{MLPChannel}(x_{out})$
\ENDFOR
\STATE \textbf{return} $(f_0, f_1, f_2, f_3)$
\end{algorithmic}
\label{alg:Encoder}
\end{algorithm}

Note that the Mamba blocks only modulate the magnitude spectrum $F_{mag}$, while the phase spectrum $F_{pha}$ is explicitly preserved and re-integrated during the inverse transformation. This design follows a decoupled representation learning strategy: the magnitude captures global contextual energy, while the phase serves as a structural anchor to maintain the spatial topology. Unlike spatial Mamba which selects discrete voxels, the Frequency Mamba acts as an adaptive spectral filter. In contrast to Fourier-based convolutions such as Fast Fourier Convolution (FFC)~\cite{chi2020fast} that employ static kernels, SSFMamba modulates the state-space matrices $\mathbf{B}$ and $\mathbf{C}$ conditioned on the magnitude spectrum $F_{mag}$. This mechanism effectively generates a data-dependent transfer function, allowing the model to dynamically prioritize low-frequency structural coherence or high-frequency edge refinements based on the specific anatomical characteristics of each scan. Such adaptive filtering overcomes the static limitations of traditional spectral methods, providing a robust, learnable mechanism to focus on the most informative spectral bands.

\begin{algorithm}[!tb]
\caption{SSFMamba Decoder (Up-sampling Block)}
\begin{algorithmic}[1]
\REQUIRE 
Input feature map $x$, skip connection feature map $s$
\ENSURE 
Up-sampled feature map $y$
\STATE $x = \mathrm{TransposedConv}(x;$
\STATE \hspace{1.5em} $\text{out\_channels}=C_{out},\ \text{kernel}=2,\ \text{stride}=2)$
\STATE $x = \mathrm{Concat}(x, s;\ \text{dim}=1)$
\IF{use residual block}
    \STATE $y = \mathrm{UnetResBlock}(x;\ \text{kernel}=3,\ \text{norm}=instance)$
\ELSE
    \STATE $y = \mathrm{UnetBasicBlock}(x;$
    \STATE \hspace{1.5em}  $\ \text{kernel}=3,\ \text{norm}=instance)$
\ENDIF
\STATE \textbf{return} $y$
\end{algorithmic}
\label{alg:Decoder}
\end{algorithm}

\textbf{Spatial-Domain Branch.} 
For each spatial-domain branch, input feature \( F_{i, j} \) is normalized through layer normalization and then fed into the Mamba module to extract features. These features are subsequently unified in both dimensionality and channels with the frequency-domain information.

{\small
\begin{equation}
    F_{spa}=\mathrm{Mamba}(\mathrm{LayerNorm}(F_{i, j})).
\end{equation}
}

\textbf{Multi-domain Information Fusion.} After both branches have extracted spatial and frequency-domain features, respectively, the frequency-domain image is converted back to the spatial domain using the inverse Fourier transform. The multi-directional frequency-domain images are then aligned, and their features are fused and enhanced through element-wise addition. An MLP module is employed to achieve superior fusion performance.

{\small
\begin{equation}
\begin{gathered}
    F_{fus} = \mathrm{MLP}(F_{spa} + F_{fre_1} + F_{fre_2} + F_{fre_3}),\\
    F_{i, j}^{out} = F_{i, j} + F_{fus}.
\end{gathered}
\end{equation}
}

The \(\mathrm{MLP}\) module comprises two 1$\times$1$\times$1 3D convolutions with a GELU nonlinearity in between. Finally, a residual connection adds the input features to the processed output, and the result is forwarded to the subsequent module.

\subsubsection{\textbf{3D Multi-Directional Scanning Mechanism}}\label{sec:mdsm} The original Mamba sequence scanning is designed for modeling unidirectional sequences, making it less suitable for image processing, especially for high-dimensional 3D medical images. For 2D images, scanning mechanisms along multiple directions have been proposed, such as horizontal and vertical traversals in both forward and reverse orders \cite{huang2024localmamba}. For 3D image scanning, a 2D scanning mechanism is applied to each individual slice, and the slices are then concatenated sequentially. Our scanning mechanism takes full advantage of the inherent properties of 3D images as well as the corresponding relationships in frequency-domain features. 
We adopt three scanning orders, namely in-slice, cross-slice, and 3D local ordering, whose selection is motivated by the distinct neighborhood structures inherent in 3D medical volumes. In-slice ordering prioritizes planar adjacency within each acquisition plane, capturing intra-slice anatomical continuity. Cross-slice ordering positions voxels from adjacent slices closer in the sequence, thereby exposing through-plane correspondences that are often overlooked by single-direction scans. 3D local ordering groups voxels within a local cube before advancing to the next, preserving isotropic spatial proximity. Together, these three paths cover complementary anatomical relationships—planar, inter-slice, and volumetric—ensuring that the resulting spectral sequences capture a comprehensive representation of the underlying 3D structure.

Since image data are real-valued, their Fourier spectra exhibit Hermitian symmetry (Eq.~\eqref{eq:hermitian}). 
As discussed in Section~\ref{sec:freq}, for the raster-type cross-slice and in-slice orderings, the serialized magnitude sequence is approximately palindromic, so that a single forward scan captures both forward and reverse spectral contexts. The 3D local ordering does not preserve this mirrored structure; instead, it complements the raster orderings by keeping spectrally adjacent components close in the sequence. The scanning mechanism is illustrated in Fig.~\ref{fig:MDSM}. 
Since fusion is performed after restoring all directional features to the same spatial coordinate system, the operation aggregates aligned anatomical responses rather than averaging misaligned features, thereby reducing the risk of localized boundary blurring.
In this way, the MDSM strengthens global spectral modeling, while each ordering retains a specific neighborhood structure (planar, through-plane, or volumetric) within its sequence.

\begin{algorithm}[!tb]
\caption{SSFMamba Model Forward}
\begin{algorithmic}[1]
\REQUIRE Input tensor $x$
\ENSURE Output tensor (e.g., segmentation)
\STATE $outs = \text{SSFMambaEncoder}(x)$ \COMMENT{Multi-scale features}
\STATE $enc1 = \text{UnetrBasicBlock}(x)$ \COMMENT{Initial encoder block}
\STATE $d5 = \text{UpBlock5}(outs[3], outs[2])$
\STATE $d4 = \text{UpBlock4}(d5, outs[1])$
\STATE $d3 = \text{UpBlock3}(d4, outs[0])$
\STATE $d2 = \text{UpBlock2}(d3, enc1)$
\STATE $dec = \text{UpBlock1}(d2)$
\STATE \textbf{return} $\text{UnetOutBlock}(dec)$
\end{algorithmic}
\label{alg:SSFMamba}
\end{algorithm}

\subsubsection{\textbf{Decoder Block}} 
As illustrated in Fig.~\ref{fig:SSFMamba}, transposed convolutions are initially employed to upsample the spatial dimensions of the lower module's feature maps. Subsequently, these upsampled feature maps are concatenated with the corresponding skip connection feature maps from the encoder, and two convolutional layers are utilized to achieve feature fusion and restore high-resolution representations. Additionally, residual connections are incorporated to facilitate effective deep-network training. The pseudo-code of the decoder block is shown in Algorithm~\ref{alg:Decoder}. Finally, the pseudo-code of the overall architecture is presented in Algorithm~\ref{alg:SSFMamba}, where skip connections are employed to bridge the encoder and decoder, facilitating feature fusion across different scales and improving segmentation performance.

\subsection{Loss Function} 
We formulate 3D medical image segmentation as a voxel-wise multi-class classification problem, where each voxel is assigned to one of $K$ categories. For the BraTS datasets, $K=4$, covering the necrotic tumor core (NCR), peritumoral edema (ED), enhancing tumor (ET), and background; for the BTCV dataset, $K=14$, covering 13 abdominal organs and background. Following this formulation, we employ the cross-entropy loss to optimize the network by measuring the divergence between the predicted class probability distribution and the ground-truth labels.

The cross-entropy loss~\cite{mao2023cross, feng2021can} jointly integrates softmax normalization and the negative log-likelihood, enabling stable optimization and effective gradient propagation during training. By explicitly modeling the competition among multiple classes, it encourages confident predictions for the correct class while suppressing ambiguous or incorrect responses. This property is particularly beneficial for dense voxel-level prediction tasks, as it provides probabilistic outputs and facilitates convergence in multi-class segmentation settings. The loss function is defined as:

{\small
\begin{equation}
L = -\frac{1}{N}\sum_{i=1}^{N} \log\left(\frac{e^{z_{y_i}}}{\sum_{j=1}^{K} e^{z_j}}\right),
\end{equation}
}

where $N$ denotes the total number of voxels, $z$ represents the predicted logits for each class, and $z_{y_i}$ corresponds to the logit associated with the ground-truth class of voxel $i$. For the BTCV dataset, following common practice in multi-organ segmentation, a Dice loss term is additionally combined with the cross-entropy loss to alleviate the severe class imbalance among organs of different sizes.

\section{Experiments}

\subsection{Datasets and implementation} 
\subsubsection{Datasets} 
We evaluate our proposed approach with the BraTS2023 \cite{menze2014multimodal, baid2021rsna} and BraTS2020 \cite{bakas2017advancing, bakas2018identifying} 
datasets. 
The BraTS2020 and BraTS2023 datasets from the Brain Tumor Segmentation (BraTS) challenge are widely used public benchmarks for multimodal brain tumor segmentation. Both datasets consist of preoperative MRI scans of glioma patients, including four modalities: T1, contrast-enhanced T1 (T1ce), T2, and FLAIR, which provide complementary anatomical and pathological information. All images are skull-stripped, rigidly registered to a common anatomical space, resampled to an isotropic resolution of 1~mm$^3$.In addition to the official preprocessing, we further apply per-modality z-score normalization to each volume before training. Each subject is provided as a 3D volume with a typical size of $240\times240\times155$.  Expert annotations delineate different tumor subregions, including necrotic and non-enhancing tumor core, peritumoral edema, and enhancing tumor. Following the standard BraTS evaluation protocol, these labels are grouped into three clinically relevant regions: whole tumor (WT), tumor core (TC), and enhancing tumor (ET). Compared with BraTS2020, BraTS2023 includes a larger and more diverse multi-center cohort with improved annotation consistency and more challenging tumor appearances, enabling a more comprehensive evaluation of model robustness and generalization ability.

In addition to the BraTS datasets, we further evaluate our method on the BTCV (Beyond the Cranial Vault) dataset~\cite{landman2015miccai}, which is a widely used benchmark for multi-organ abdominal segmentation. The dataset consists of contrast-enhanced abdominal CT scans acquired from multiple clinical centers. Each volume is annotated by experienced radiologists with voxel-wise labels for major abdominal organs, including the liver, spleen, kidneys, pancreas, stomach, and other anatomical structures.
All CT volumes are provided in 3D format with varying spatial resolutions and organ appearances, posing significant challenges due to large inter-organ shape variations and low-contrast boundaries. The BTCV dataset is commonly used in the MICCAI multi-organ segmentation challenge and serves as an important benchmark for evaluating the generalization capability of segmentation models beyond brain imaging tasks~\cite{wasserthal2023totalsegmentator}.

\subsubsection{Implementation details} 
Our model was implemented using Python 3.10.13 and PyTorch 2.1.1 with NVIDIA GeForce RTX 3090 GPUs and CUDA 11.8. All 3D medical images were randomly cropped to dimensions of 128$\times$128$\times$128 for model training, with a batch size set to 2. 
We employed the SGD optimizer to update the network parameters, setting the learning rate to $10^{-2}$ with a weight decay of $3\times10^{-5}$. 
A high momentum coefficient of 0.99 together with Nesterov acceleration was further adopted to promote smoother parameter updates and facilitate stable convergence during training. Meanwhile, prior to entering the Mamba blocks, features are normalized through the network normalization layers, resulting in standardized feature distributions that improve optimization stability.
The datasets were divided into a training set (80\%) and a test set (20\%). For the BTCV dataset, we adopted a 60\%/40\% split (18 cases for training and 12 for testing), with the learning rate set to $10^{-3}$. These adjustments only account for the much smaller size of BTCV, and the network architecture remains unchanged across all datasets.

\subsection{Comparison with State-of-the-art Methods} 
We compare our proposed SSFMamba with several state-of-the-art segmentation methods, including CNN-based approaches (U-Net \cite{ronneberger2015u}, UX-Net \cite{lee2023uxnet}, MedNeXt \cite{roy2023mednext}, SuperLightNet \cite{yu2025superlightnet}), Transformer-based approaches (UNETR \cite{hatamizadeh2022unetr}, SwinUNETR \cite{hatamizadeh2022swinunetr}, SwinUNETR-V2 \cite{he2023swinunetr}, TransUNet \cite{chen2024transunet}), and Mamba-based approaches (SegMamba \cite{xing2024segmamba}, EM-Net \cite{chang2024net}). To ensure a fair comparison, we retrained these models under the same settings and data processing procedures.

\begin{table*}[!tb]
\centering
\small
\renewcommand{\arraystretch}{1.35}
\begin{tabular}{r|c|cccccccc}
\hline
\multirow{2}{*}{Methods} & \multirow{2}{*}{Architecture}  & \multicolumn{2}{c}{WT} & \multicolumn{2}{c}{TC}  & \multicolumn{2}{c}{ET}   & \multicolumn{2}{c}{Avg} \\
&  & Dice$\uparrow$ & HD95$\downarrow$ & Dice$\uparrow$ & HD95$\downarrow$ & Dice$\uparrow$ & HD95$\downarrow$ & Dice$\uparrow$ & HD95$\downarrow$ \\ \hline
U-Net 2015 & &92.83  &4.17  &92.15   &3.76    &86.75 & \multicolumn{1}{l|}{4.20} &90.58 &4.05 \\
UX-Net 2023  & &\underline{94.55} &3.69  &92.63 &4.29  &86.98 & \multicolumn{1}{l|}{3.86} &91.39 &3.95 \\
MedNeXt 2023 &&93.94 &3.57  &91.42 &3.68  &88.41 & \multicolumn{1}{l|}{3.92} &91.26 &3.72 \\ 
SuperLightNet 2025 & \multirow{-4}{*}{CNN}&94.42&3.94&\underline{93.76}&3.28&88.61 & \multicolumn{1}{l|}{3.74} &92.26&3.65 \\ \cline{1-2}
UNETR 2022&    &93.30 &4.88  &90.02 &5.88  &86.56 & \multicolumn{1}{l|}{5.35} &89.96 &5.37  \\
SwinUNETR 2022 &\multirow{2}{*}{Transformer} &93.77&4.10&90.14&4.45&87.07 & \multicolumn{1}{l|}{4.38} &90.33&4.31 \\
SwinUNETR-V2 2023 & &93.57 &4.03  &92.89 &4.38  &88.97 & \multicolumn{1}{l|}{3.92} &91.81 &4.11 \\
TransUNet 2024 &  &92.44 &4.31  &87.57 &4.41  &81.65 &  \multicolumn{1}{l|}{4.73} &87.22 &4.48 \\ \cline{1-2}
EM-Net 2024 & &93.04 & \underline{3.53} &92.53 &3.79  &88.84 & \multicolumn{1}{l|}{4.03} &91.47 &3.78  \\
SegMamba 2024 & \multirow{-2}{*}{Mamba}&93.57 & 3.69 &93.55 &\underline{3.04}  &\underline{89.76} & \multicolumn{1}{l|}{\textbf{3.31}} &\underline{92.30} &\underline{3.35}\\ \hline
SSFMamba (Ours) & Mamba & \textbf{94.69} & \textbf{3.41}& \textbf{94.78} & \textbf{2.96} & \textbf{89.92}& \multicolumn{1}{l|}{\underline{3.43}} & \textbf{93.13}&\textbf{3.26} \\ \hline
\end{tabular}
\vspace{2mm}
\caption{Quantitative comparison on the BraTS2023 dataset. The best and second-best results are highlighted in bold and underlined.}
\label{tab:2023}
\vspace{-1.5mm}
\end{table*}

\begin{table*}[!tb]
\centering
\small
\renewcommand{\arraystretch}{1.35}
\begin{tabular}{r|c|llllllll}
\hline
\multirow{2}{*}{Methods} & \multirow{2}{*}{Architecture}  & \multicolumn{2}{c}{WT} & \multicolumn{2}{c}{TC} & \multicolumn{2}{c}{ET}& \multicolumn{2}{c}{Avg} \\
& \multicolumn{1}{l|}{}& Dice$\uparrow$& HD95$\downarrow$ & Dice$\uparrow$& HD95$\downarrow$ & Dice$\uparrow$ & HD95$\downarrow$  & Dice$\uparrow$& HD95$\downarrow$\\ \hline
U-Net 2015 &   &89.79&4.91&84.01&5.56&74.82 & \multicolumn{1}{l|}{8.01} &82.87&6.16 \\  
UX-Net 2023 &&90.09&4.62&83.32&5.60&76.19 & \multicolumn{1}{l|}{8.06} &83.20&6.09 \\
MedNeXt 2023 &&\underline{91.26}&4.75&83.52&\textbf{4.77}&75.71 & \multicolumn{1}{l|}{8.01} &83.49&\underline{5.84} \\
SuperLightNet 2025 & \multirow{-4}{*}{CNN}&90.68&7.89&83.16&5.65&\underline{76.31} & \multicolumn{1}{l|}{7.71} &83.38&7.08 \\ \cline{1-2}
UNETR 2022 & &90.13&6.05&78.93&6.11&73.89 & \multicolumn{1}{l|}{8.72} &80.98&6.96 \\
SwinUNETR 2022 &\multirow{2}{*}{Transformer} &90.10&5.13&84.11&5.54&75.52 & \multicolumn{1}{l|}{7.91} &83.24&6.19 \\
SwinUNETR-V2 2023 & &90.37&5.16&83.83&5.40&75.18 & \multicolumn{1}{l|}{\underline{7.58}} &83.13&6.04 \\ 
TransUNet 2024 & &89.35  &\underline{4.34}&80.34&5.67&71.47 & \multicolumn{1}{l|}{8.64} &80.38 &6.22  \\ \cline{1-2}
EM-Net 2024 &&90.81&5.04&83.44&5.77&75.65 & \multicolumn{1}{l|}{8.08} &83.30&6.29 \\
SegMamba 2024 & \multirow{-2}{*}{Mamba} &90.77&4.38&\underline{84.33}&6.17&75.98 & \multicolumn{1}{l|}{8.04} &\underline{83.69}&6.20 \\
\hline
SSFMamba (Ours)  & Mamba &\textbf{91.31}&\textbf{4.31}&\textbf{84.51}&\underline{4.97}&\textbf{76.69} & \multicolumn{1}{l|}{\textbf{7.53}} &\textbf{84.17} &\textbf{5.60} \\ \hline
\end{tabular}
\vspace{2mm}
\caption{Quantitative comparison on the BraTS2020 dataset. The best and second-best results are highlighted in bold and underlined.}
\label{tab:2020}
\vspace{-1.5mm}
\end{table*}

\begin{table*}[!tb]
\centering
\small
\setlength{\tabcolsep}{3pt}
\renewcommand{\arraystretch}{1.65}
\begin{tabular}{r|c|ccccccccccccc|c}
\hline
Methods & Architecture 
& Spl & RKid & LKid & Gall & Eso & Liv & Sto & Aor & IVC & Psv & Pan & RAG & LAG & Avg. \\
\hline
U-Net 2015        
& \multirow{4}{*}{CNN}
&\underline{92.48}  &91.23  &85.92  &59.53  &\underline{75.77}  &95.63  &80.07  &89.39  &84.13  &69.63  &62.88  &65.81  &66.59  &78.39  \\
UX-Net 2023        
&                       
&91.33  &91.41  &\underline{90.37}  &53.17  &74.09  &95.42  &78.08  &89.85  &85.19  &69.78  &67.64  &67.47  &63.93  &78.29  \\
MedNeXt 2023      
&                       
& 92.16 & \underline{92.31} & 86.27 &66.14  &74.31  &95.75  &82.10  &89.81  &\underline{86.31}  &\textbf{74.29}  &69.97 &69.09  &66.06  &80.37  \\
SuperLightNet 2025 
&
&85.42  &85.13  &86.82  &\textbf{69.15}  &75.25  &93.85  &76.89  &89.13  &85.54  &70.28  &70.51  &67.18  &67.93  &78.70  \\
\hline
UNETR 2022        
& \multirow{4}{*}{Transformer}
&83.52  &80.35  &78.25  &53.97  &70.46  &93.54  &71.39  &86.84  &78.91  &59.69  &60.32  &62.33  &59.78  &72.26  \\
SwinUNETR 2022 
&                       
&86.28  &83.50  &75.39  &50.23  &70.44  &94.16  &77.18  &85.32  &78.25  &59.26  &53.55  &56.12  &52.49  &70.94  \\
SwinUNETR-V2 2023 
&                       
&83.81  &75.61  &78.47  &49.95  &71.86  &95.82  &68.07  &89.86  &82.17  &66.96  &48.12  &57.50  &49.70  &70.61  \\
TransUNet 2024    
&                       
&86.43  &83.73  &76.55  &55.61  &69.89  &94.21  &77.82  &85.05  &77.90  &59.28  &68.32  &63.76  &53.84  &73.26  \\
\hline
EM-Net 2024       
& \multirow{2}{*}{Mamba}
&92.43 &91.57 &90.34 &\underline{67.65}  &75.13  &95.84  &\textbf{83.98}  &\textbf{90.66}  &\textbf{86.47}  &73.03  &\underline{74.98}  &\textbf{70.10}  &\underline{68.09}  &\underline{81.55}  \\
SegMamba 2024     
&                       
&92.32  &89.95  &85.78  &65.74  &73.22  &\underline{95.95}  &83.04  &89.24  &86.22  &72.60  &73.62  &65.68  &67.28  &80.06  \\
\hline
SSFMamba (Ours)   
& Mamba                 
& \textbf{95.44} & \textbf{94.34} & \textbf{94.07} & 62.53 & \textbf{76.12} & \textbf{96.67} & \underline{83.93} & \underline{89.94} & 84.27 & \underline{74.19} & \textbf{81.97} & \underline{69.30} & \textbf{68.83} & \textbf{82.43} \\
\hline
\end{tabular}
\vspace{2mm}
\caption{Quantitative comparison on the BTCV dataset (Dice, \%). Spl: spleen; RKid/LKid: right/left kidney; Gall: gallbladder; Eso: esophagus; Liv: liver; Sto: stomach; Aor: aorta; IVC: inferior vena cava; Psv: portal and splenic veins; Pan: pancreas; RAG/LAG: right/left adrenal gland. The best and second-best results are highlighted in bold and underlined.}
\label{tab:btc}
\vspace{1.5mm}
\end{table*}

As presented in Table~\ref{tab:2023} and Table~\ref{tab:2020}, 
we can observe that the proposed approach achieves the best average Dice and average HD95 on both datasets. On the BraTS2023 dataset—which features a larger and more diverse set of cases—our method ranks first in five of the six subregion metrics and second in the remaining one (HD95 of ET). In terms of HD95, the incorporation of frequency-domain features enhances global contextual understanding, leading to better boundary alignment. Our approach achieves a 0.83\% improvement in average Dice over the second-best method (SegMamba), demonstrating its robustness and effectiveness in handling complex segmentation tasks. 

Benefiting from the state-space-based modeling in SSFMamba, which enables efficient global dependency modeling and stable feature propagation, our method achieves the highest Dice for all three tumor subregions on the BraTS2020 dataset, together with the lowest HD95 for WT and ET.
Compared with existing CNN, Transformer, and Mamba-based approaches, SSFMamba obtains the highest average Dice (84.17\%) and the lowest average HD95 (5.60), indicating improved global consistency and more accurate boundary delineation.

To further evaluate the generalization capability of our approach, we extend the model to the BTCV dataset, which consists of single-modality CT images for multi-organ segmentation. As presented in Table~\ref{tab:btc}, our method achieves the highest average Dice score on the BTCV dataset, outperforming the second-best method by 0.88\%, which indicates its strong generalization ability across imaging modalities and anatomical structures. In particular, SSFMamba achieves 81.97\% Dice on the pancreas, surpassing the second-best method (EM-Net, 74.98\%) by 6.99\%. This result is particularly noteworthy as the pancreas is traditionally considered a ``hard-to-segment'' organ in spatial-driven methods due to its ill-defined boundaries and high morphological variability. Nevertheless, SSFMamba does not achieve the best results on several small structures with large shape variability, such as the gallbladder and the inferior vena cava, suggesting that there is still room for improvement on such organs.

\begin{figure*}[!tb]
  \centering

\begin{minipage}{\textwidth}
  \centering
  \includegraphics[width=0.95\textwidth, angle=0]{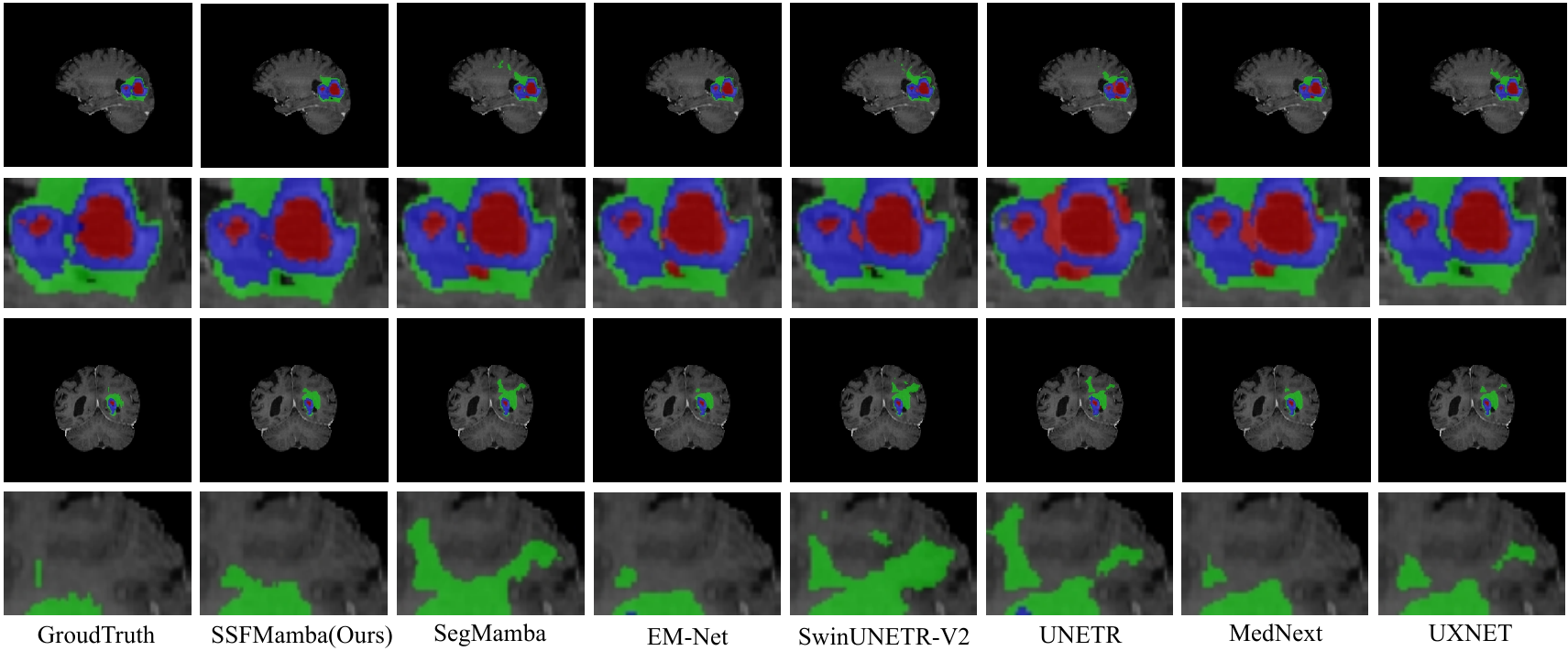}
  \vspace{-2mm}
  \caption{Qualitative comparison on the BraTS2023 dataset. The annotated regions are categorized into three classes: red indicates NCR (necrotic tumor core), blue indicates ET (Gd-enhancing tumor), and green indicates ED (the peritumoral edematous/invaded tissue). Below each image, a magnified view of the corresponding region is provided to highlight the differences in fine-grained segmentation performance across methods.}
  \label{fig:cmp}
\end{minipage}
  \vspace{6mm}

\begin{minipage}{\textwidth}
  \centering
  \includegraphics[width=0.95\textwidth, angle=0]{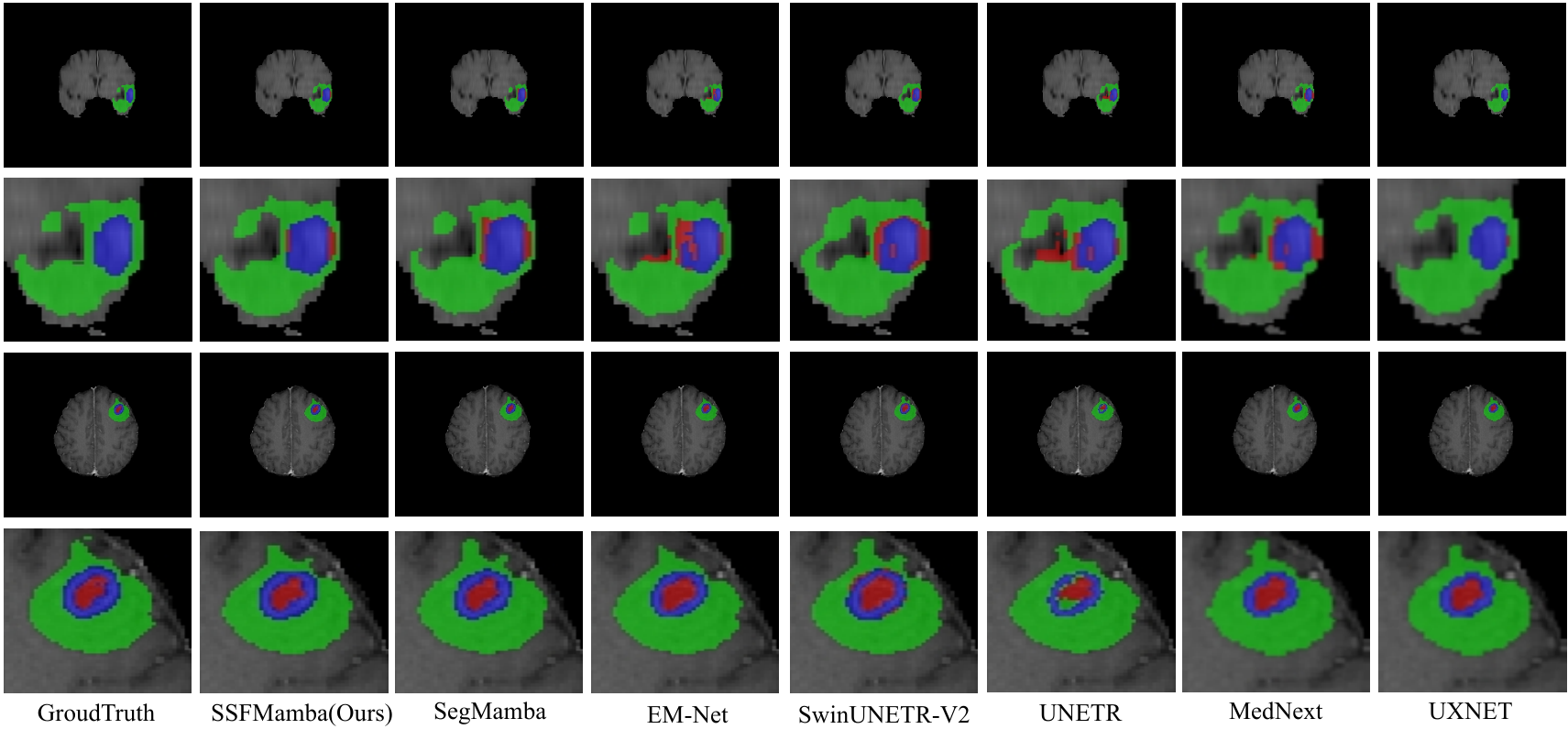}
  \vspace{-2mm}
  \caption{Qualitative comparison on the BraTS2020 dataset. The annotated regions are categorized into three classes: red indicates NCR (necrotic tumor core), blue indicates ET (Gd-enhancing tumor), and green indicates ED (the peritumoral edematous/invaded tissue). Below each image, a magnified view of the corresponding region is provided to highlight the differences in fine-grained segmentation performance across methods.}
  \label{fig:cmp_seg_2}
\end{minipage}

\end{figure*}

\begin{figure*}[!htbp]
  \centering
  \includegraphics[width=0.95\textwidth, angle=0]{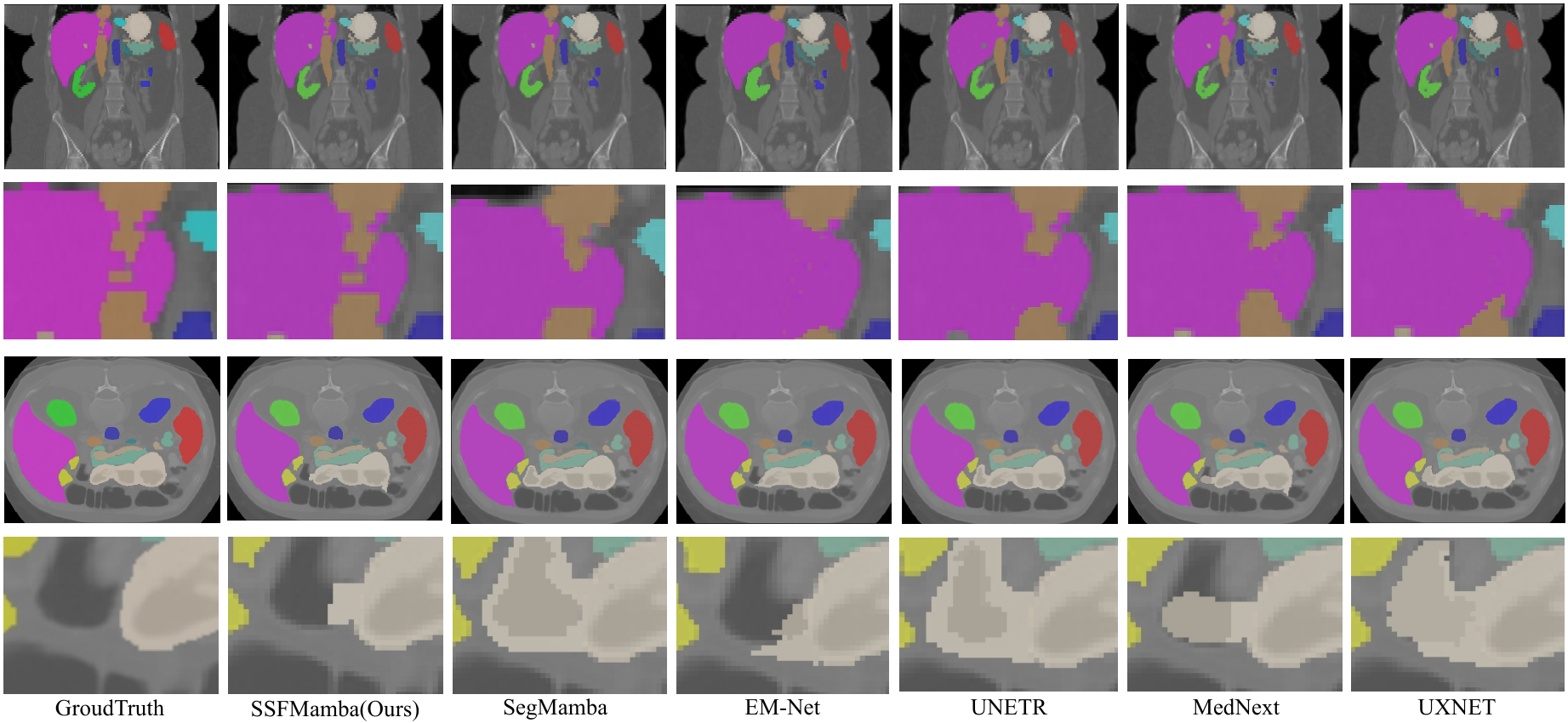}
  \vspace{-2mm}
  \caption{Qualitative comparison on the BTCV dataset. The annotated regions delineate multiple abdominal organs, including the liver, spleen, kidneys, pancreas, and stomach. Below each image, a magnified view of the corresponding region is provided to highlight the differences in fine-grained segmentation performance across methods.}
  \label{fig:cmp_btcv_2}
  \vspace{2mm}
\end{figure*}

To more intuitively validate the superior performance of our method, we further visualize the segmentation results of the various methods on the dataset, as illustrated in Fig.~\ref{fig:cmp}, Fig.~\ref{fig:cmp_seg_2} and Fig.~\ref{fig:cmp_btcv_2}. We can observe that the compared methods are more likely to over-segment or under-segment, especially on subtle edges. In contrast, even in complex segmentation scenarios, our approach delineates the boundaries of various labels more clearly. Additionally, in terms of local details, our method achieves more refined segmentation. 

The feature maps in Fig.~\ref{fig:comparison}(a)--(d) further confirm the complementary roles of the two branches: the spatial branch yields more complete semantic representations, the frequency branch highlights high-contrast contours, and their fusion preserves both semantic completeness and boundary sharpness.

\textbf{Empirical Validation of Modality Independence.} The superior average performance of SSFMamba on both MRI (BraTS) and CT (BTCV) benchmarks provides empirical support for the modality-invariant hypothesis proposed in Section~\ref{sec:intro}. Notably, SSFMamba achieves this without any architecture modification across these heterogeneous datasets; only the learning rate and data split are adjusted to accommodate the much smaller size of BTCV. Traditional spatial features are sensitive to domain shifts arising from varied scanner protocols or organ contrasts. In contrast, the spectral domain offers a more stable representation: low-frequency components describing global organ envelopes remain consistent across modalities and provide a stable structural prior, while mid-to-high frequency components encode tissue interfaces. The benefit is most prominent for the low-contrast pancreas, where SSFMamba improves the Dice score by 6.99\% over the second-best method. We attribute this gain to the adaptive spectral filtering of the frequency branch: built upon the stable low-frequency structure, it can emphasize the subtle mid-to-high frequency components of tissue interfaces that are blurred in the spatial domain due to limited intensity contrast. These results suggest that spectral selectivity is an effective tool for capturing heterogeneous anatomical details across diverse clinical scenarios.

\begin{table*}[!tb]
\centering
\small
\renewcommand{\arraystretch}{1.4}
\begin{tabular}{p{1.05cm}<{\centering}p{1.05cm}<{\centering}p{1.05cm}<{\centering}p{1.05cm}<{\centering}|p{1.55cm}<{\centering}|cccccccc}
\hline
\multirow{2}{*}{Frequency}&\multirow{2}{*}{Cross-slice}&\multirow{2}{*}{In-slice}&\multirow{2}{*}{3D Local}&\multirow{2}{*}{Architecture}& \multicolumn{2}{c}{WT} & \multicolumn{2}{c}{TC} & \multicolumn{2}{c}{ET}    & \multicolumn{2}{c}{Avg} \\
&&&&& Dice$\uparrow$& HD95$\downarrow$ & Dice$\uparrow$& HD95$\downarrow$ & Dice$\uparrow$ & HD95$\downarrow$  & Dice$\uparrow$& HD95$\downarrow$\\ \hline
&\checkmark&&&&93.33 &3.91 &92.77    &3.40 &88.09& \multicolumn{1}{l|}{3.87}  &91.40    &3.73 \\
&\checkmark&\checkmark&& &92.79 &3.83 &92.92    &3.45 &89.46& \multicolumn{1}{l|}{\textbf{3.29}}  &91.73    &3.52 \\
&\checkmark&\checkmark&\checkmark& &\underline{94.31} &3.59 &92.00    &3.09 &89.23&  \multicolumn{1}{l|}{3.61}  &91.85    &3.43 \\
\checkmark&\checkmark &\ & & &93.68 &\textbf{3.40} &92.79    &3.04 &\textbf{90.96}& \multicolumn{1}{l|}{3.66}  &\underline{92.46}    &3.37 \\ 
\checkmark& &\checkmark & &Mamba &93.65 & 4.12 &93.32    & 3.80 & 88.69 & \multicolumn{1}{l|}{4.03}  & 91.89    &3.98 \\ 
\checkmark& & &\checkmark & &94.26 &3.53 &93.06    &3.06 & 88.43 & \multicolumn{1}{l|}{3.51}  &91.92    &3.36 \\ 
\checkmark&\checkmark &\checkmark & & &\underline{94.31} &3.56 &91.82    &3.43 &89.58& \multicolumn{1}{l|}{3.60}  &91.90   &3.53 \\ 
\checkmark&\checkmark & &\checkmark & &93.75 & 3.45 &\underline{93.81}    &\underline{2.97} & 89.49 & \multicolumn{1}{l|}{3.55}  & 92.35    & \underline{3.33} \\ 
\checkmark& &\checkmark &\checkmark & &93.83 &3.55 &92.69    &3.06 &89.77& \multicolumn{1}{l|}{3.71}  &92.09    &3.44 \\ \hline

\checkmark&\checkmark&\checkmark&\checkmark & CNN  &92.60 &4.39 &90.21    &4.69 &85.96& \multicolumn{1}{l|}{4.24}  &89.59    &4.44 \\
\checkmark&\checkmark&\checkmark&\checkmark &Transformer &93.23 &4.02 &92.41    &3.55 &87.73& \multicolumn{1}{l|}{3.96}  &91.12    &3.84 \\
\hline
\checkmark&\checkmark&\checkmark&\checkmark&Mamba &\textbf{94.69} &\underline{3.41}&\textbf{94.78}    &\textbf{2.96}&\underline{89.92}& \multicolumn{1}{l|}{\underline{3.43}}  &\textbf{93.13}    &\textbf{3.26} \\ \hline
\end{tabular}
\vspace{1.5mm}
\caption{Ablation study for the effectiveness of each component on the BraTS2023 dataset.}
\label{tab:Ablation1}
\vspace{-5mm}
\end{table*}

\subsection{Ablation Analysis} 
We conduct ablation studies on the BraTS2023 dataset to assess the contributions of various components within our overall model. 
As presented in Table~\ref{tab:Ablation1}, the first three rows remove the frequency branch and apply the scanning orderings only to the spatial-domain Mamba. Compared with the full model, these variants show a clear performance drop (91.40\%--91.85\% vs. 93.13\% average Dice), primarily because the absence of frequency-domain information weakens the model's ability to capture global context. 
In the spatial domain, progressively adding the in-slice and 3D local orderings yields steady but modest gains (91.40\% $\rightarrow$ 91.73\% $\rightarrow$ 91.85\%), since spatial-domain data do not possess the Hermitian symmetry exploited by the MDSM. In the frequency domain, any single ordering already achieves 91.89\%--92.46\% average Dice, and combining all three orderings yields the best result (93.13\%), whereas pairwise combinations do not consistently outperform the best single ordering. This indicates that the three orderings are complementary and that their full combination is required to fully exploit the spectral structure.

In addition, we replace the Mamba backbone with CNN and Transformer counterparts while keeping all other components, and both variants underperform our Mamba-based design (89.59\% and 91.12\% vs. 93.13\% average Dice). This is because the MDSM is specifically tailored for the sequence-sensitive nature of Mamba: the scanning orderings determine how spectral components interact in the causal recurrence, a property that has no direct counterpart in convolution- or attention-based backbones. Each MDIF Encoder incorporates two MDIF Blocks. For comparison, we also evaluate encoder configurations containing one and three MDIF Blocks (Table~\ref{tab:Ablation2}). Quantitative evaluations show that our design with two MDIF Blocks delivers superior performance. These ablation experiments verify the rationale and effectiveness of our designed modules.

\begin{table*}[!tb]
\centering
\small
\renewcommand{\arraystretch}{1.4}
\begin{tabular}{l|cccccccc}
\hline
\multirow{2}{*}{Methods}& \multicolumn{2}{c}{WT} & \multicolumn{2}{c}{TC} & \multicolumn{2}{c}{ET}    & \multicolumn{2}{c}{Avg} \\
& Dice$\uparrow$& HD95$\downarrow$ & Dice$\uparrow$& HD95$\downarrow$ & Dice$\uparrow$ & HD95$\downarrow$  & Dice$\uparrow$& HD95$\downarrow$\\ \hline
1 MDIF Block &\underline{94.21} &\underline{3.59}&92.26    &\underline{3.14}&88.25& \multicolumn{1}{l|}{3.58}  &91.57    &\underline{3.44} \\ 

2 MDIF Blocks (SSFMamba)            &\textbf{94.69} &\textbf{3.41}&\textbf{94.78}    &\textbf{2.96}&\textbf{89.92}& \multicolumn{1}{l|}{\textbf{3.43}}  &\textbf{93.13}    &\textbf{3.26} \\ 

3 MDIF Blocks &92.83 &4.23&\underline{93.53}    &3.44&\underline{89.72}& \multicolumn{1}{l|}{\underline{3.57}}  &\underline{92.03}    &3.75 \\ \hline
\end{tabular}
\vspace{2mm}
\caption{Comparison of different numbers of MDIF Blocks per encoder stage on the BraTS2023 dataset.}
\label{tab:Ablation2}
\end{table*}

\section{Conclusion}
In this paper, we presented SSFMamba, a novel symmetry-driven spatial-frequency fusion framework for 3D medical image segmentation. 
By integrating the 3D MDSM with SSMs, our approach effectively alleviates the directional biases inherent in traditional 1D scanning and enables more balanced global modeling of causal sequences. 
The Frequency Mamba in the frequency branch facilitates adaptive spectral modeling, enabling the network to emphasize informative anatomical features across different frequency components.
Extensive experiments on the BraTS2020, BraTS2023, and BTCV datasets demonstrate that SSFMamba achieves the best average performance compared with state-of-the-art methods in both MRI and CT modalities. The remarkable performance on low-contrast abdominal organs—exemplified by the 81.97\% Dice on the pancreas—suggests that our framework effectively captures subtle spectral footprints often obscured in the spatial domain. These results support the effectiveness of exploiting intrinsic physical consistency within the frequency domain. 
By leveraging spectral representations that capture intrinsic anatomical structures beyond modality-specific appearance patterns, SSFMamba fosters modality-agnostic representations and yields consistent performance across imaging modalities without architecture modification. Future work will explore cross-modality transfer, computational efficiency analysis, and its applicability to broader medical imaging scenarios.

\bibliographystyle{IEEEtran}
\bibliography{refs}


 





\end{document}